\documentclass[runningheads]{llncs}
\usepackage{graphicx}
\usepackage{subcaption}

\usepackage{tikz}
\usetikzlibrary{calc, positioning, arrows.meta, matrix}
\usepackage{comment}
\usepackage{amsmath,amssymb} 
\usepackage{color}

\makeatletter
\renewcommand*{\@fnsymbol}[1]{\ensuremath{\ifcase#1\or *\or \dagger\or \ddagger\or
    \mathsection\or \mathparagraph\or \|\or **\or \dagger\dagger
    \or \ddagger\ddagger \else\@ctrerr\fi}}
\makeatother
\usepackage[colorlinks,linkcolor={black},citecolor={green},urlcolor={black},backref=page]{hyperref}

\usepackage{booktabs}

\usepackage[width=122mm,left=12mm,paperwidth=146mm,height=193mm,top=12mm,paperheight=217mm]{geometry}

\begin{document}
\pagestyle{headings}
\mainmatter
\def\ECCVSubNumber{23}  

\title{Can we cover navigational perception needs of the visually impaired by panoptic segmentation?}


\titlerunning{Can we cover perception needs by panoptic segmentation?}
%
\author
{Wei Mao\thanks{Equal Contribution}\inst{1,2},
Jiaming Zhang  \inst{*1},
Kailun Yang\thanks{Correspondence: kailun.yang@kit.edu}\inst{1},
Rainer Stiefelhagen  \inst{1} 
}
%
%
\authorrunning{W. Mao et al.}
%
\institute{Karlsruhe Institute of Technology, Karlsruhe 76131, Germany \and
Tongji University, Shanghai 200092, China \\
}
\maketitle

\begin{abstract}
Navigational perception for visually impaired people has been substantially promoted by both classic and deep learning based segmentation methods.
In classic visual recognition methods, the segmentation models are mostly object-dependent, which means a specific algorithm has to be devised for the object of interest.
In contrast, deep learning based models such as instance segmentation and semantic segmentation allow to individually recognize part of the entire scene, namely \textit{things} or \textit{stuff}, for blind individuals.
However, both of them can not provide a holistic understanding of the surroundings  for the visually impaired.
Panoptic segmentation is a newly proposed visual model with the aim of unifying semantic segmentation and instance segmentation.
Motivated by that, we propose to utilize panoptic segmentation as an approach to navigating visually impaired people by offering both things and stuff awareness in the proximity of the visually impaired. We demonstrate that panoptic segmentation is able to equip the visually impaired with a holistic real-world scene perception through a wearable assistive system. 

\end{abstract}

\section{Introduction}
\label{sec:intro}
According to the world health organization, at least 2.2 billion people have a vision impairment or
blindness~\cite{bourne2017magnitude}.
The majority of people with visual impairments still use simple and conventional assistive tools,
e.g., white canes, which is able to sense the nearby objects in front of the person but can not provide any information about the categories of the objects, let alone background understanding such as the localization of road areas and buildings~\cite{schwarze2015intuitive}.
The development of navigational assistance aiming to help the visually impaired to reach their destinations safely and independently becomes increasingly challenging, as it requires detecting a wide variety of scene elements to provide higher-level assistive awareness via scene understanding~\cite{yang2018unifying}.

To date, navigational perception for the visually impaired is still primarily founded on the computer vision techniques, either classic computer vision algorithms or modern deep learning based ones. The early approaches~\cite{cheng2018real,rodriguez2012assisting,schwarze2015intuitive,wang2017enabling} mainly attached great importance to the so-called generic object detection where the generic objects range from foreground obstacles to traversable space without considering the semantics for concrete segments~\cite{watson2020footprints}.
These navigation tasks for blind people majorly took advantage of the classic image segmentation methods such as region growing, connected components labeling and random field without any semantic interpretation being involved~\cite{yang2016expanding}. 
Taking into account a limited level of semantics of the environment, for example, a pedestrian detection system with wearable cameras was designed for detecting the nearby pedestrians for blind people~\cite{lee2020pedestrian}.

\begin{figure}[t]
    \centering
    \includegraphics[width=0.9\linewidth]{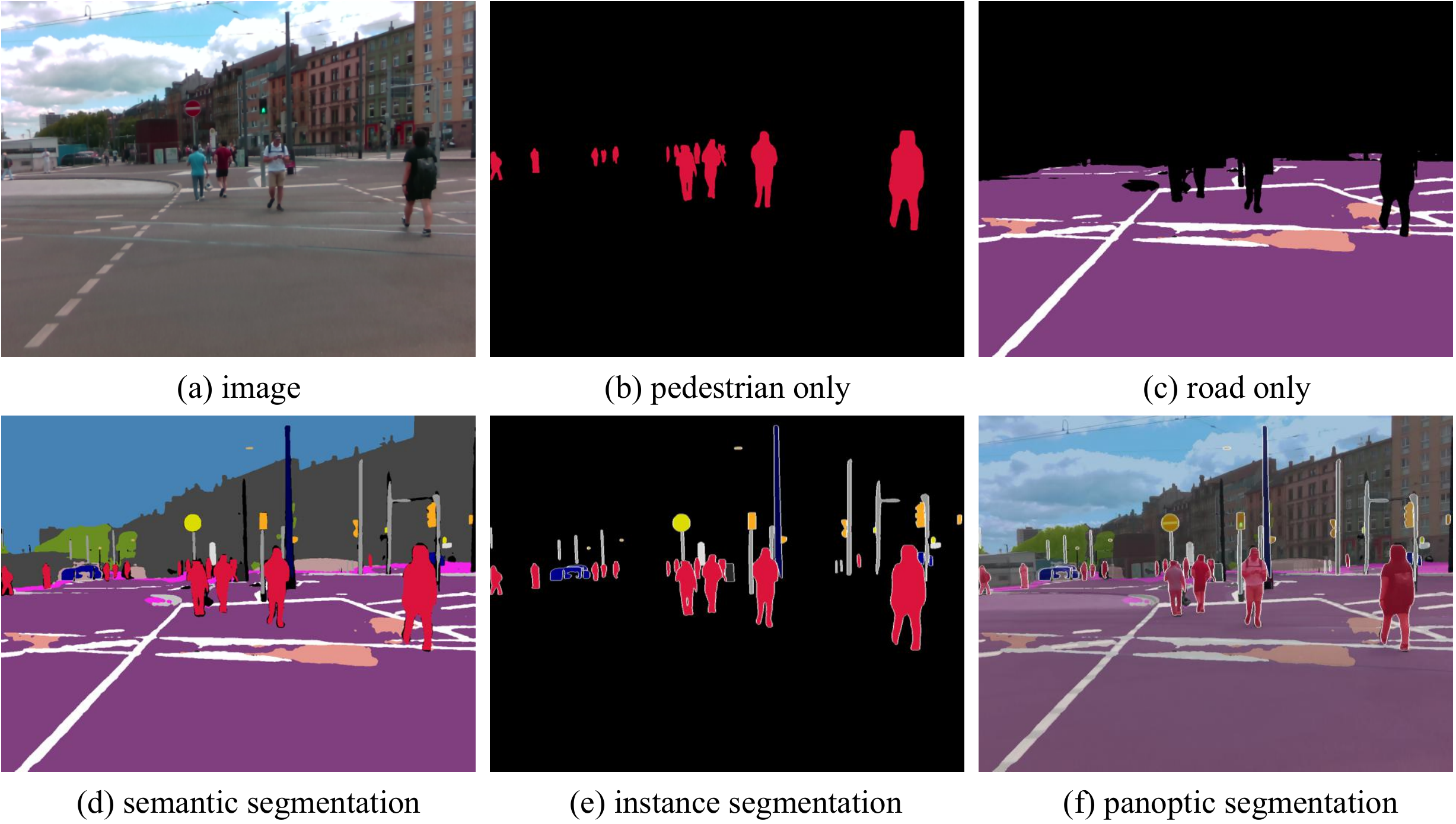}
    \caption{Contrast between different levels of scene perception for the input image (a): (b) only pedestrians are recognized, (c) only roadways are identified, (d) semantic segmentation without differentiating instance of the same category, (e) instance segmentation with ignorance about the background stuff, (f) panoptic segmentation by unifying semantic and instance segmentation.}
    \label{fig:concept}
\end{figure}

The aforementioned methods can assist people with visual impairments by enabling the detection of limited kinds of objects in the environment. Nevertheless, the obvious downside of them is that we have to exploit different models for detection of different objects, which adds to the complications for navigational system deployment~\cite{long2019assisting}. In the real world, visually impaired people constantly encounter various sorts of objects, especially in outdoor environment. Hence, they must be informed about what exact kinds of objects they are facing towards narrowing the gap of environment perception between blind and sighted people. However, it is quite challenging to cope with this problem merely based on the classic computer vision methods.

Recent years witness considerable advancements in deep learning based computer vision techniques such as object detection~\cite{ren2016faster,tian2019fcos}, semantic segmentation~\cite{chen2017deeplab,chen2018encoder,long2015fully,zhao2017pyramid} and instance segmentation~\cite{he2017mask,neven2019instance,novotny2018semi,wang2019solo}.
Such compelling techniques enable the development of many frameworks with more comprehensive and efficient perception for the visually impaired.
For instance, the popular instance segmentation framework Mask R-CNN~\cite{he2017mask} has been adopted to aid visually impaired people to identify a wide spectrum of objects which may occur in their surroundings~\cite{long2019unifying,yohannes2019content}.
Apart from the countable objects like pedestrians, vehicles and such, collectively termed things, the stuff, amorphous regions of similar texture or material such as road, sky and grass~\cite{kirillov2019panoptic}, should also be incorporated into the visual assistance task on the grounds that the identification of traffic infrastructure like road and sidewalk is essential for the visually impaired to participate in traffic safely.
As an example of stuff perception for the blind people, a light-weight semantic segmentation network was proposed to simultaneously address the detection of blind road and crosswalk specifically~\cite{cao2020rapid}, but overlooking the thing classes like various obstacles in the scene.

Naturally, it leads us to the question: \textit{how to unify the recognition tasks for the typically distinct thing classes and stuff classes in the neighborhood of the visually impaired people?} Fortunately, a recently proposed novel visual recognition task called panoptic segmentation answers this question perfectly in that it unifies instance and semantic segmentation problems by mapping each pixel in a image to a pair of class label and instance ID~\cite{kirillov2019panoptic}, which serve as comprehensive visual perception information communicated to the visually impaired for securing their participation in the traffic.

For this reason, in this paper we intend to present how the recently emerging panoptic segmentation can be employed to help navigate visually impaired people by providing both stuff and thing interpretations in a single network model, alongside a wearable assistive system. 

\section{Related Work}
\label{sec:related_work}
\textbf{Classical Visual Assistance.} In the early assistance system for visually impaired people via scene understanding, some researchers have attempted to extract scene information with a stereo camera system.
Rodriguez et al.~\cite{rodriguez2012assisting} segmented the image into background and obstacles based on dense disparity maps and ground plane estimation algorithms, which was extended by Yang et al.~\cite{yang2016expanding} to support long-range traversable area detection.
Schwarze et al.~\cite{schwarze2015intuitive} also adopted the classic stereo vision methods and offered more information about the scene with obstacle detection and tracking.
Nonetheless, these approaches do not distinguish specific instances such as pedestrians, cars, bikes etc.

\noindent\textbf{Semantic Segmentation.} Semantic segmentation aims at predicting a per-pixel class label without making any distinction among multiple instances belonging to the same category. State-of-the-art approaches to semantic segmentation adopt the widely used Fully Convolutional Network (FCN)~\cite{long2015fully} as its basic skeleton. Since the contextual information benefits the semantic prediction for pixels, the dilated convolutions were often integrated into the FCN~\cite{chen2017deeplab,chen2018encoder} in order to exploit the context with larger receptive field. Furthermore, spatial pyramid pooling was also used to extract multi-scale contextual information for better performance~\cite{zhao2017pyramid}. However, semantic segmentation can not tell apart the instances of the same category.

\noindent\textbf{Instance Segmentation.} Instance segmentation attaches the bulk of attention to the instance recognition in contrast with semantic segmentation. Evolving from the object detection, instance segmentation predicts class probabilities, bounding boxes and instance masks in a image. As the name indicates, instance segmentation identifies distinct instances of the same category. Research community has put forward various approaches to instance segmentation. These methods can be split into two categories: two-stage methods and one-stage methods. In two-stage methods, a group of region proposals is generated first and then passed to a sub-network for classification and segmentation, e.g, Mask R-CNN~\cite{he2017mask}. Although two-stage methods accomplish state-of-the-art accuracy, they are infeasible to use for real-time applications owing to the high computational costs. Conversely, one-stage method achieves faster inference while keeping relatively qualified accuracy, e.g., SOLO~\cite{wang2019solo}. Despite the good trade-off between speed and accuracy that has been achieved, instance segmentation disregards the background stuff intrinsically.

\noindent\textbf{Segmentation for Assistance.}
Recent assistance approaches for the visually impaired emerge with the renaissance of the deep learning~\cite{duh2020v}.
Yohannes et al.~\cite{yohannes2019content} applied the notable instance segmentation framework Mask R-CNN~\cite{he2017mask} to recognize various objects in the outdoor environment in place of the traditional models  but neglecting stuff classes including road and sidewalk. 
Yang et al.~\cite{yang2018intersection} leveraged real-time semantic segmentation to coordinate the perception needs of visually impaired pedestrians at traffic intersections, including the detection of crosswalk and pedestrian crossing lights.
Hua et al.~\cite{hua2019small} proposed a RGB-D semantic segmentation system that is sensitive to small obstacles, which allows to negotiate low-lying obstacles that pose significant challenges during navigation.
Additionally, Cao et al.~\cite{cao2020rapid} developed a light-weight semantic
segmentation network for fast detection of blind road and sidewalk without making any distinction between multiple instances of the same category inherently.

\noindent\textbf{Panoptic Segmentation.}
Unlike previous methods, we propose to cover navigational perception needs of visually impaired pedestrians by using panoptic segmentation. 
Panoptic segmentation, as defined by~\cite{kirillov2019panoptic}, is able to assign one class label for each pixel in an image and an instance ID for all pixels belonging to an object, i.e. a high level of scene understanding.
In the context of navigating the visually impaired, panoptic segmentation unifies the semantic and instance segmentation by taking into account both foreground things and background stuff.
That is exactly what we need for navigating the visually impaired with a comprehensive interpretation of the images. 
Depending on the architecture, the panoptic segmentation methods are categorized into two classes: two-net methods and one-net methods. In two-net methods, two separate networks for semantic and instance segmentation, respectively, are trained and their outputs are merged into the panoptic output format. This method can achieve top accuracy at the cost of speed~\cite{wang2019joint}. By comparison, one-net method is more advantageous in terms of inference speed while keeping decent accuracy for diverse kinds of application, e.g. Panoptic FPN~\cite{kirillov2019pfpn} and seamless segmentation~\cite{porzi2019seamless}. To the best of our knowledge, we are the first to utilize panoptic segmentation to assist the navigational perception for visually impaired people.

\section{System}
\subsection{Wearable Navigation System for the Visually Impaired}

\begin{figure}[h!]
    \centering
    \includegraphics[width=0.8\columnwidth]{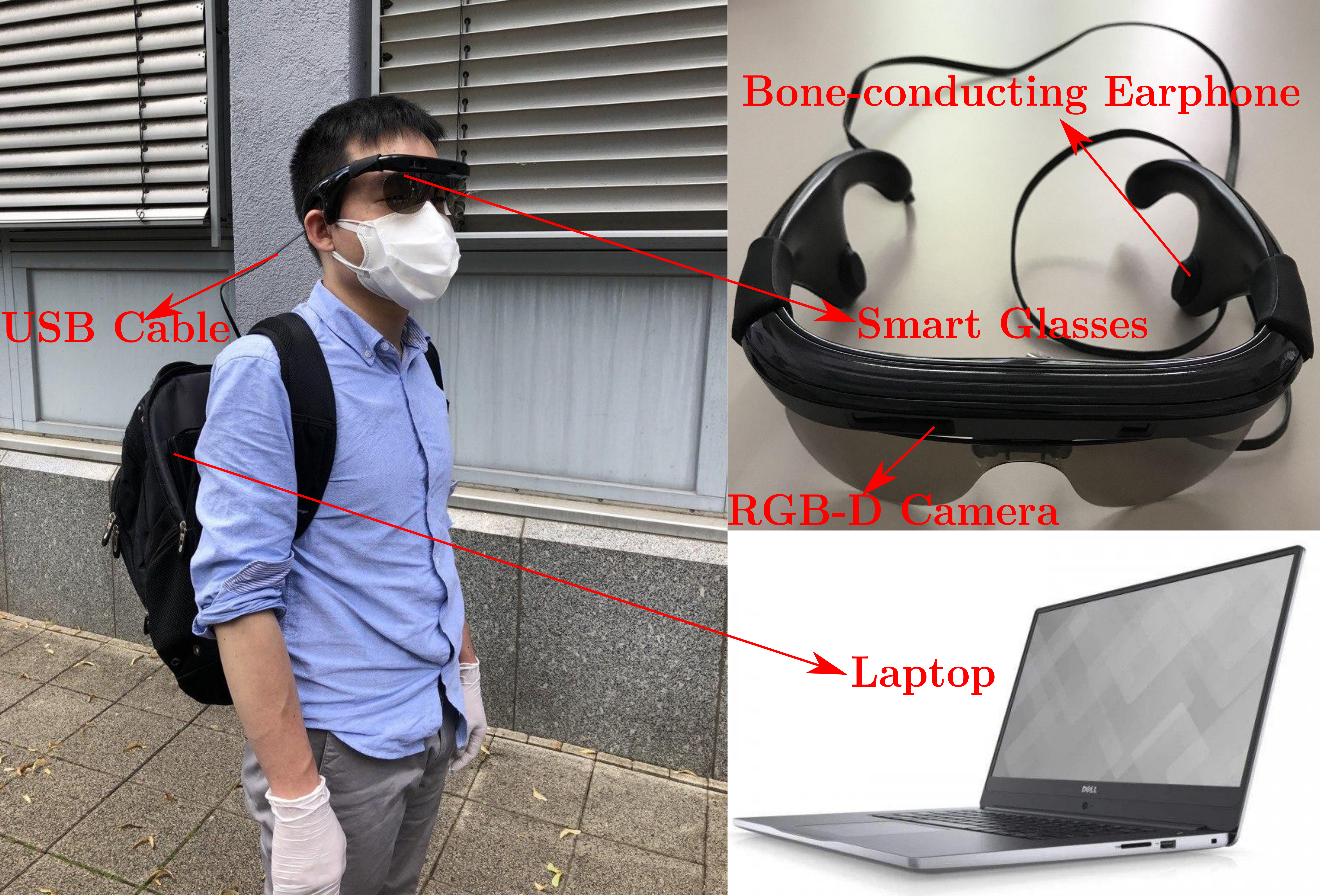}
    \caption{Illustration of the wearable assistive system for visually impaired people. The entire system (left pane) consists of two parts connected by an USB cable: a pair of smart glasses with a RGB-D camera as its integral part (upper right) and a laptop with sufficient computation power for CNNs calculation (lower right).}
    \label{fig:wearable_sysl}
\end{figure}

In order to deploy the panoptic segmentation framework to assist the visually impaired, we design a compact wearable assistive system as illustrated in the Figure~\ref{fig:wearable_sysl}, which is comprised of a pair of smart glasses with a RGB-D camera RealSense R200~\cite{keselman2017intel} embedded and a laptop working as a computing node. The light-weight yet powerful RGB-D camera can capture a wide variety of scenes in the outdoor environment from the perspective of visually impaired people, and  generates both RGB and depth images at the resolution of up to 1080$\times$1920 pixels and at the speed of 30 frames per second. While the visually impaired equipped with the wearable assistive system is walking outdoors, the RGB-D camera is constantly taking images of the surroundings. 

For the images captured by the RGB-D camera to be processed properly in real time, a laptop with a powerful CPU (2.7GHz Intel Core i7) and a GPU (Nvidia GeForce GTX 940MX) is chosen to make the panoptic segmentation inferences. The images are transmitted from the smart glasses to the laptop carried by the visually people on the fly via a USB cable. After panoptic segmentation is made on the received images, the panoptic perception results as shown in the Figure~\ref{fig:concept} are communicated to visually impaired people in the form of acoustic feedback through the bone-conduction earphones integrated on the glasses. The acoustic feedback can be generated by using different sonification methods such as the three real-time scene sonification methods proposed by Hu et al.~\cite{hu2020comparative} so that visually impaired people can understand the surrounding environment comprehensively. Since the main role of the laptop is to perform the deep learning calculation, we can replace it with any other computationally more efficient processing unit.

\subsection{Panoptic Segmentation Framework for the Visually Impaired}

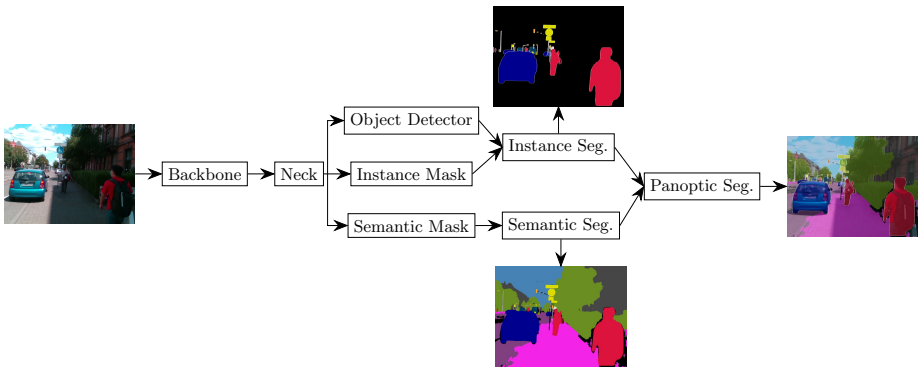
\begin{figure}[htbp!]
		\centering
		\resizebox{\columnwidth}{!}{%
			\begin{tikzpicture}[
	roundnode/.style={circle, draw=green!60, fill=green!5, very thick, minimum size=7mm},
	squarednode/.style={rectangle, draw=red!60, fill=red!5, very thick, minimum size=5mm},
	rectnode/.style={rectangle, draw, scale=2.0},
	boldarrow/.style={-{Stealth[scale=3.0]}},
	]
	        \node[rectnode]      (backbone)                              {Backbone};
	        \node[rectnode]      (neck)          [right=of backbone]     {Neck};
	        \node[rectnode]      (inst_mask)     [right=of neck]         {Instance Mask};
	        \node[rectnode]      (obj_dect)      [above=of inst_mask]    {Object Detector};
	        \node[rectnode]      (sem_mask)      [below=of inst_mask]    {Semantic Mask};
	        \node[coordinate]    (mid_obj_inst)  at ($ (obj_dect.east) !.5! (inst_mask.east) $){};
	        \node[rectnode]      (inst_seg)      [right=of mid_obj_inst] {Instance Seg.};
	        \node[rectnode]      (sem_seg)       [right=of sem_mask]     {Semantic Seg.};
	        \node[coordinate]    (mid_sem_inst)  at ($ (inst_seg.east) !.5! (sem_seg.east) $){};
	        \node[rectnode]      (pan_seg)       [right=of mid_sem_inst] {Panoptic Seg.};
	        \node[inner sep=0pt] (rgb_img) [left=of backbone]
	        {\includegraphics[width=0.4\textwidth]{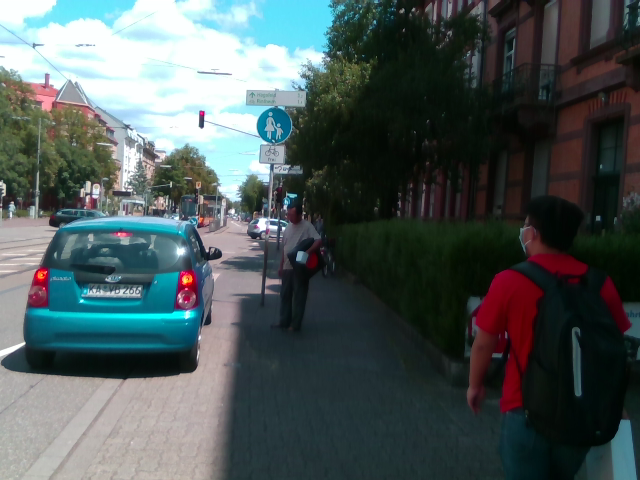}};
	        \node[inner sep=0pt] (ins_img) [above=of inst_seg]
	        {\includegraphics[width=0.4\textwidth]{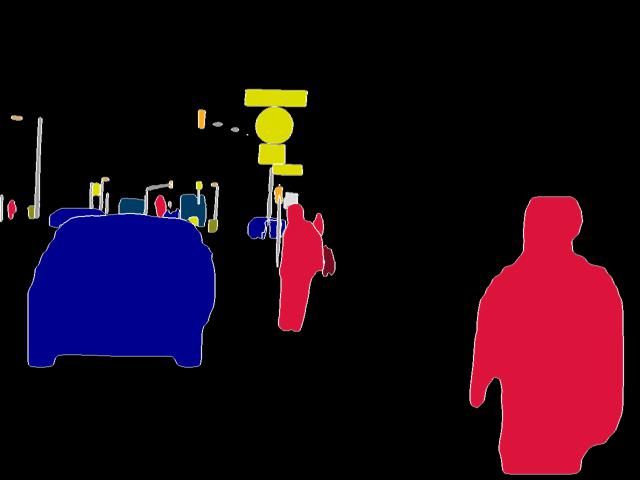}};
	        \node[inner sep=0pt] (sem_img) [below=of sem_seg]
	        {\includegraphics[width=0.4\textwidth]{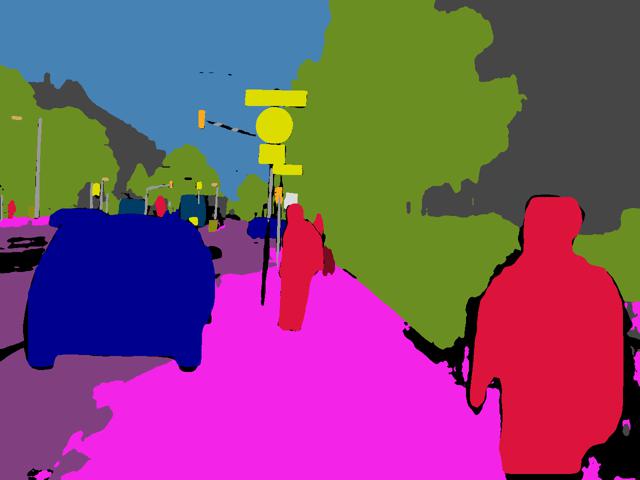}};
	        \node[inner sep=0pt] (pan_img) [right=of pan_seg]
	        {\includegraphics[width=0.4\textwidth]{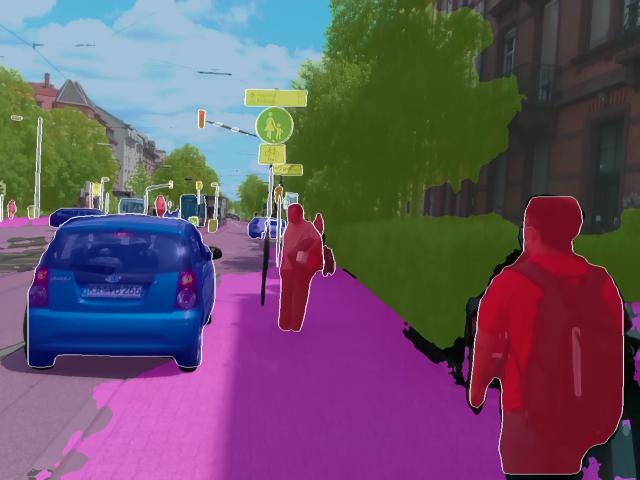}};
	        
	        \draw[boldarrow] (backbone.east) -- (neck.west);
	        \draw[boldarrow] (neck.east) -- (inst_mask.west);
	        \draw[boldarrow] ($(neck.east) + (4pt, 0)$) |- (obj_dect.west);
	        \draw[boldarrow] ($(neck.east) + (4pt, 0)$) |- (sem_mask.west);
	        \draw[boldarrow] (sem_mask.east) -- (sem_seg.west);
	        \draw[boldarrow] (sem_seg.east) -- (pan_seg.west);
	        \draw[boldarrow] (inst_seg.east) -- (pan_seg.west);
	        \draw[boldarrow] (obj_dect.east) -- (inst_seg.west);
	        \draw[boldarrow] (inst_mask.east) -- (inst_seg.west);
	        \draw[boldarrow] (rgb_img.east) -- (backbone.west);
	        \draw[boldarrow] (inst_seg.north) -- (ins_img.south);
	        \draw[boldarrow] (sem_seg.south) -- (sem_img.north);
	        \draw[boldarrow] (pan_seg.east) -- (pan_img.west);
			\end{tikzpicture}%
		}
		\caption{The whole picture of Instance, Semantic and Panoptic Segmentation. Seg. is short for Segmentation. Backbone, e.g. ResNet-50~\cite{he2016deep} or DenseNets~\cite{huang2017densely}; Neck, e.g. FPN~\cite{lin2017feature}. Instance Segmenation output derives from the object detector output and instance mask. Panoptic Segmentation result can be obtained by merging the instance segmentation output and the corresponding semantic mask.  }
		\label{fig:inst_seg_pan}
\end{figure}

There are various methods to process the images obtained by the RGB-D camera for the interpretation of  what surround the visually impaired, as discussed in the Section~\ref{sec:intro} and Section~\ref{sec:related_work}. As illustrated in Figure~\ref{fig:concept} and Figure~\ref{fig:inst_seg_pan}, the modern deep learning method is superior in terms of the level of interpretation of the images with respect to classic computer vision methods that only detect specific scene elements. Blind and visually impaired people often expect an interpretation of the surrounding analogous to the level perceived by sighted people, in which case the deep learning based segmentation approaches stand out by virtue of the more comprehensive understanding than the classic ones.

Among the family of deep learning based segmentation tasks, panoptic segmentation lends itself better to assisting the visually impaired regarding the holistic scene understanding. As it is depicted in Figure~\ref{fig:inst_seg_pan}, the panoptic segmentation can be deemed to be the fusion of instance and semantic segmentation, addressing both instance and stuff segmentation in the scene confronted during the walking of the visually impaired. With regard to the architecture, the one-net panoptic segmentation shares the backbone and neck with semantic and instance segmentation, but owns the dedicated head for the panoptic output, a pair of class label and instance ID.

In our work, we leverage the Seamless Scene Segmentation architecture~\cite{porzi2019seamless} as the panoptic segmentation network to aid the visually impaired in unified scene perception. The proposed architecture opts for a refined ResNet-50~\cite{he2016deep} and a Feature Pyramid Network (FPN)~\cite{lin2017feature} as the shared backbone and neck, respectively. Its instance segmentation head simply follows the popular Mask R-CNN~\cite{he2017mask} architecture while the semantic segmentation head is a light-weight DeepLab-like module~\cite{chen2017deeplab}. An extension to the fusion technique proposed in~\cite{kirillov2019panoptic} is applied to the instance and semantic predictions to generate a panoptic output. The network is efficient in terms of computational efforts and model sizes compared to combined, individual models. Thereby, it well suits the online application on the wearable system for visually impaired individuals.
\section{Experiments}
\subsection{Segmentation Performance: Results on Mapillary Vistas}
Mapillary Vistas~\cite{neuhold2017mapillary} is a diverse publicly available street-level imagery dataset. It consists of 25k high-resolution images, divided into three sets of 18k/2k/25k images for training, validation and testing, respectively. The street-level images are captured from all over the world from both driver and pedestrian viewpoints and span 28 stuff classes and 37 thing classes. It contains a large quantity of images taken on sidewalks and various roadways, which reasonably match the common scenes for the visually impaired, as these street-level images cover what the visually impaired are confronted with the most in the daily life. The Seamless Scene Segmentation model~\cite{porzi2019seamless} used to assist scene understanding for the visually impaired in this work is trained on the training set and evaluated on the validation set.

We evaluate the trained Seamless Scene Segmentation model on the Mapillary Vistas dataset in terms of the PQ metric~\cite{kirillov2019panoptic}. As indicated in Table~\ref{tab:metrics}, the overall panptic segmentation performance relies on the resolution of the input image in the sense that the PQ drops with the decrease of the image resolution. This is reasonable as high-resolution inputs provide fine spatial details that are relevant to the segmentation performance. However, high-resolution inputs would require higher computation complexity and incur longer inference latency (see Table~\ref{tab:ref_time}) that are disadvantageous for online applications. To achieve a good trade-off between segmentation accuracy and inference latency, we set the resolution at $480\times 640$, although our RGB-D camera can support higher resolutions. At the resolution of $480\times 640$, the panoptic prediction is rather accurate and acceptable according to the subsequent qualitative analysis for the real-world trip images. To provide finer interpretation for the visually impaired, the RGB-D camera can be set at higher resolutions.
\begin{table}[t]
\caption{Panoptic segmentation performance on Mapillary Vistas~\cite{neuhold2017mapillary} validation dataset at  different resolutions. The $AP^d$ and $AP^i$ correspond to the performance of detection and instance segmentation, respectively. The mIoU measures the performance of semantic segmentation. The $PQ^{th}$ and $PQ^{st}$ stand for  the PQ metric for  things and stuff, respectively. Additionally, the mIoUs for typical instance classes such as pole and car, are listed in the second column.}
\label{tab:metrics}
\resizebox{\textwidth}{!}{
\begin{tabular}{c|cccccc|rrr|cc|c}
\toprule
\textbf{Resolution} &
  \textbf{pole} &
  \textbf{traffic-light} &
  \textbf{person} &
  \textbf{rider} &
  \textbf{bike} &
  \textbf{car} &
  $AP^{d}$ &
  $AP^{i}$ &
  $PQ^{th}$ &
  $mIoU$ &
  $PQ^{st}$ &
  $PQ$ \\ \midrule
$240 \times 320$   & 33.4 & 46.4 & 56.3 & 46.8 & 40.3 & 72.2 & 3.1  & 2.5  & 7.7  & 23.9 & 13.4 & 10.2 \\
$480 \times 640$   & 43.5 & 61.1 & 68.8 & 57.2 & 55.0 & 85.3 & 7.7  & 6.5  & 17.0 & 41.3 & 30.3 & 22.7 \\
$960 \times 1280$  & 54.5 & 73.0 & 78.2 & 68.3 & 66.7 & 90.6 & 13.9 & 12.5 & 28.0 & 50.6 & 41.9 & 34.0 \\
$1920 \times 2560$ & 57.8 & 76.0 & 77.4 & 63.3 & 68.2 & 92.0 & 17.6 & 16.2 & 33.6 & 52.0 & 44.4 & 38.3 \\ \bottomrule
\end{tabular}}
\end{table}

Besides, we report the mean Intersection-over-Union (mIoU) for semantic segmentation, Average Precision (AP) for instance segmentation and per-class IoU results for typical objects commonly confronted by visually impaired pedestrians such as car and person. Specifically, cars, bikes, riders and persons are dynamic participants that visually impaired people need to avoid during navigation. Poles are static obstacles, while traffic-lights are key elements at traffic intersections when visually impaired people have to cross the roads~\cite{cheng2018real}. Generally, the segmentation is more accurate for large and frequent classes available in the training set such as cars, while it is less robust for thin objects (like poles) or less frequent classes (like riders). Overall, for these safety-critical classes, while the segmentation is not perfect, it is pretty accurate at close range where objects are larger in the image, which qualifies the usage of panoptic segmentation for providing reliable assistance to the visually impaired.

In addition to the accuracy measured by the PQ metric, the inference time is also of great concern for the deployment of the wearable assitive system with panoptic segmentation capability. Hence, we compare the inference time on two different GPUs for input images at different resolutions: $240 \times 320$, $480 \times 640$, $ 960 \times 1280$. As it is shown in the Table~\ref{tab:ref_time}, the inference time for one input image diminishes considerably as the input image resolution deteriorates. For instance, it takes one GTX 1070 GPU $0.162s$ to complete the panoptic segmentation on a $480 \times 640$ RGB image. For the purpose of navigating the visually impaired, such inference speed is sufficient in view of the relatively low walking speed of the visually impaired.
\begin{table}[h]
\caption{Inference time (second per frame).}
\centering
\begin{tabular}{|c||c|c|}
    \hline
    {\textbf{Resolution}}&{\textbf{GTX 940MX (s)}}&{\textbf{GTX 1070 (s)}}\\
    \hline
    {$240 \times 320$}&{0.829}&{0.114}\\
    \hline
    {$480 \times 640$}&{1.312}&{0.162}\\
    \hline
    {$960 \times 1280$}&{2.261}&{0.259}\\
    \hline
\end{tabular}
\label{tab:ref_time}
\end{table}
\subsection{Train Here, Navigate There: Results on a Real-world Trip}
	\begin{figure}[!ht]
		\centering
		\begin{subfigure}[b]{0.475\textwidth}   
			\centering 
			\includegraphics[width=\textwidth]{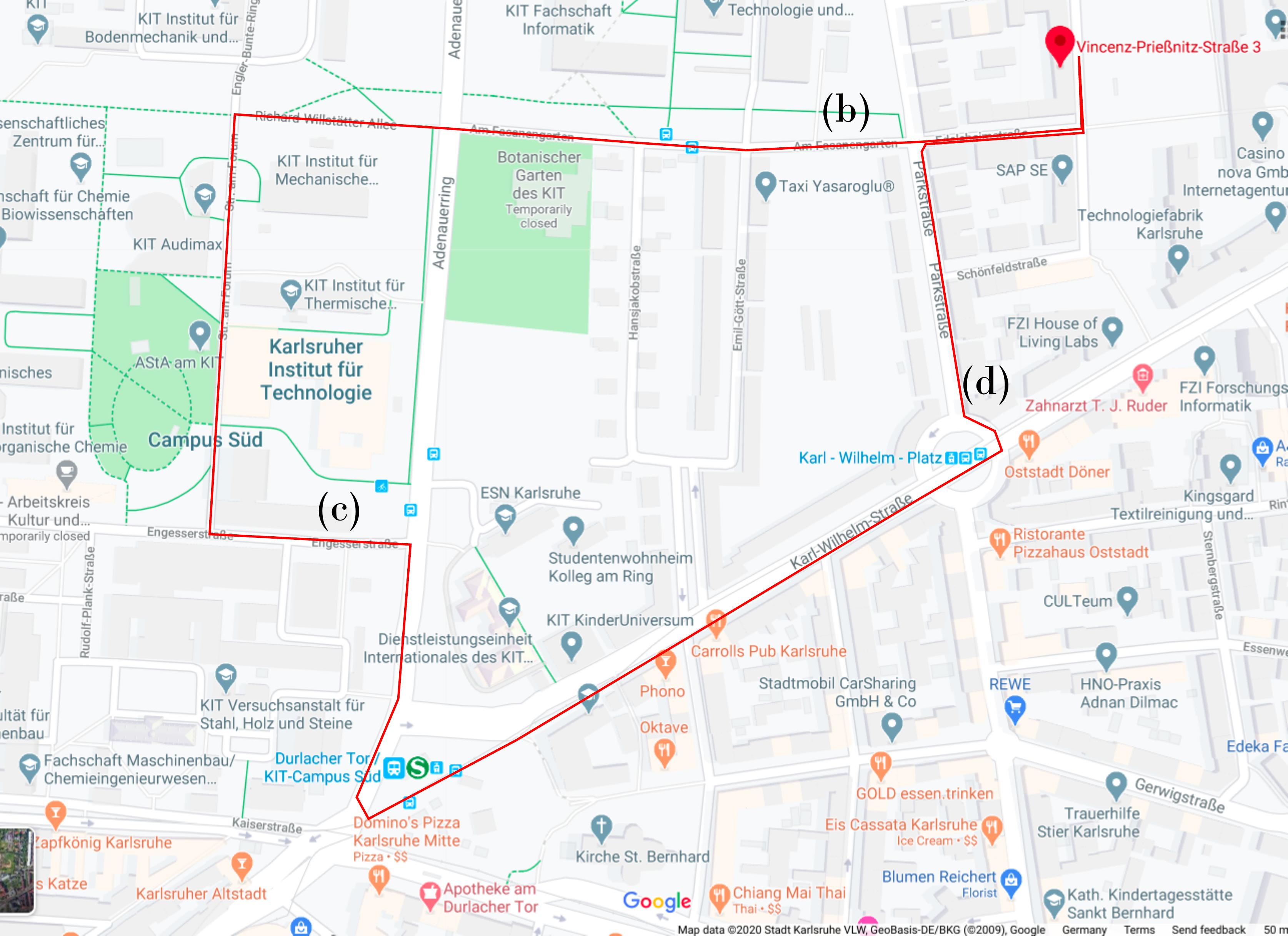}
			\caption[]%
			{{\small Walking Route}}    
			\label{fig:distr_sub_route}
		\end{subfigure}
		\hfill
		\begin{subfigure}[b]{0.475\textwidth}
			\centering
			\includegraphics[width=\textwidth]{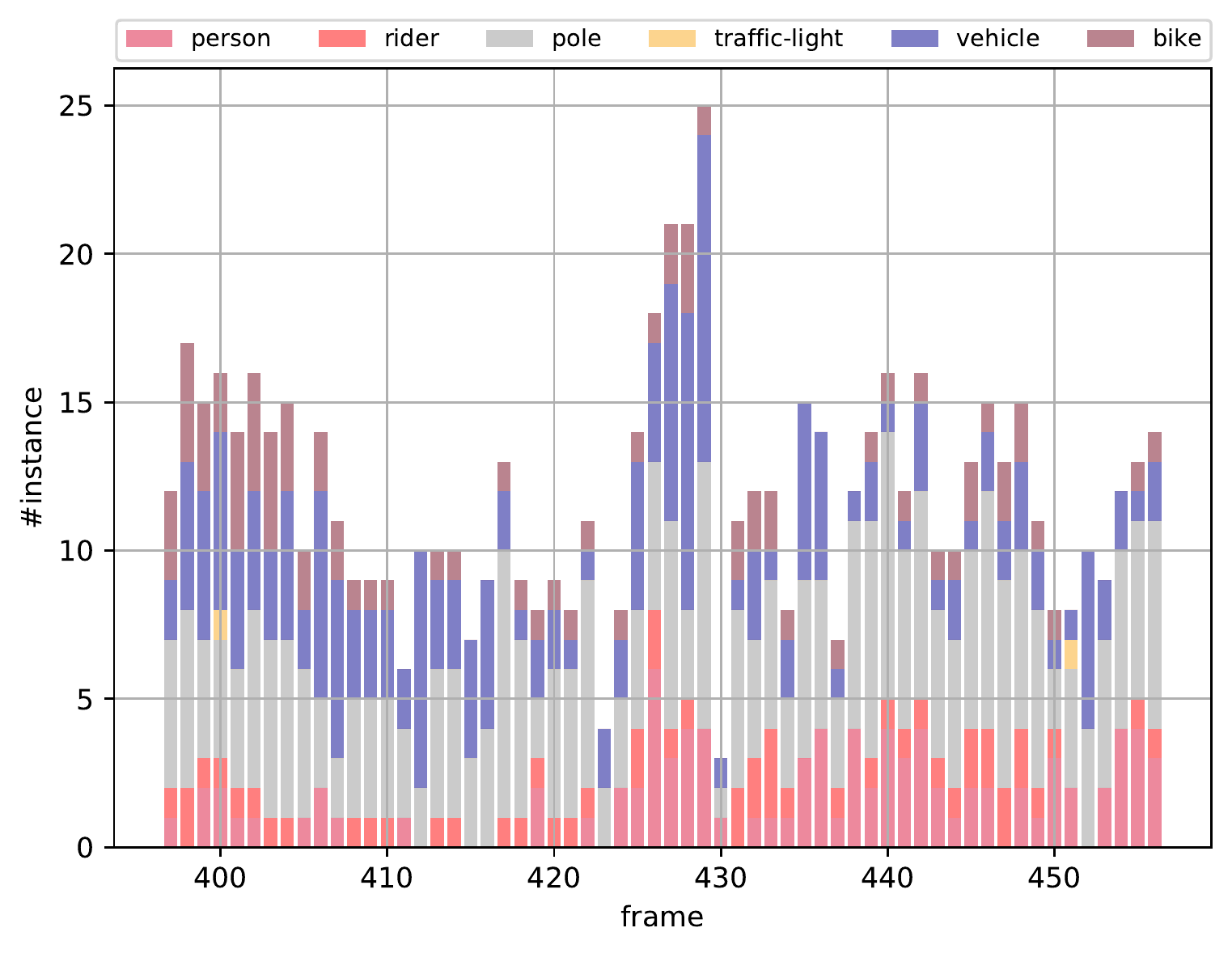}
			\caption[]%
			{{\small Begin of the Trip}}    
			\label{fig:distr_sub_begin}
		\end{subfigure}
		\vskip\baselineskip
		\begin{subfigure}[b]{0.475\textwidth}  
			\centering 
			\includegraphics[width=\textwidth]{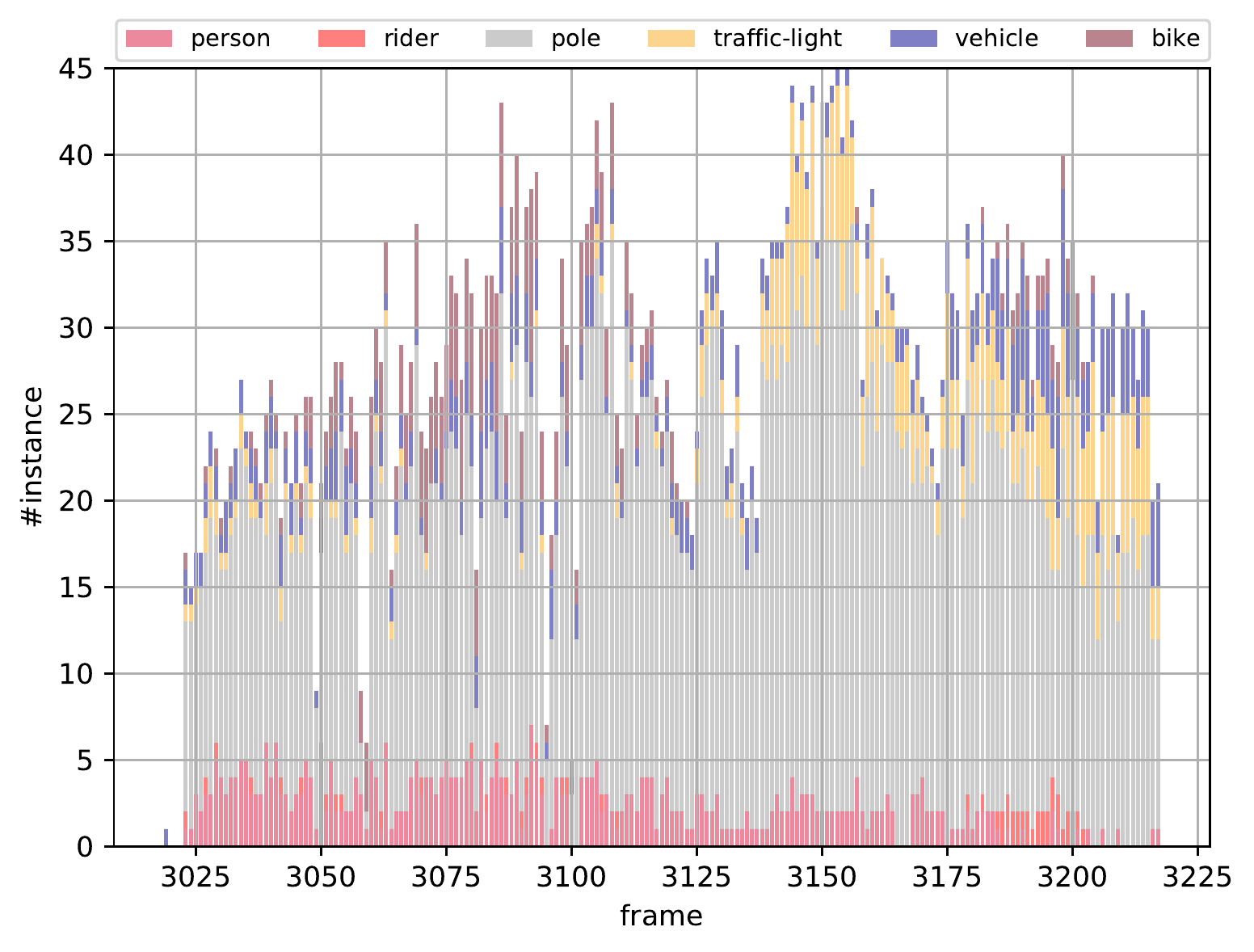}
			\caption[]%
			{{\small Halfway through the Trip}}    
			\label{fig:distr_sub_halfway}
		\end{subfigure}
		\hfill
		\begin{subfigure}[b]{0.475\textwidth}   
			\centering 
			\includegraphics[width=\textwidth]{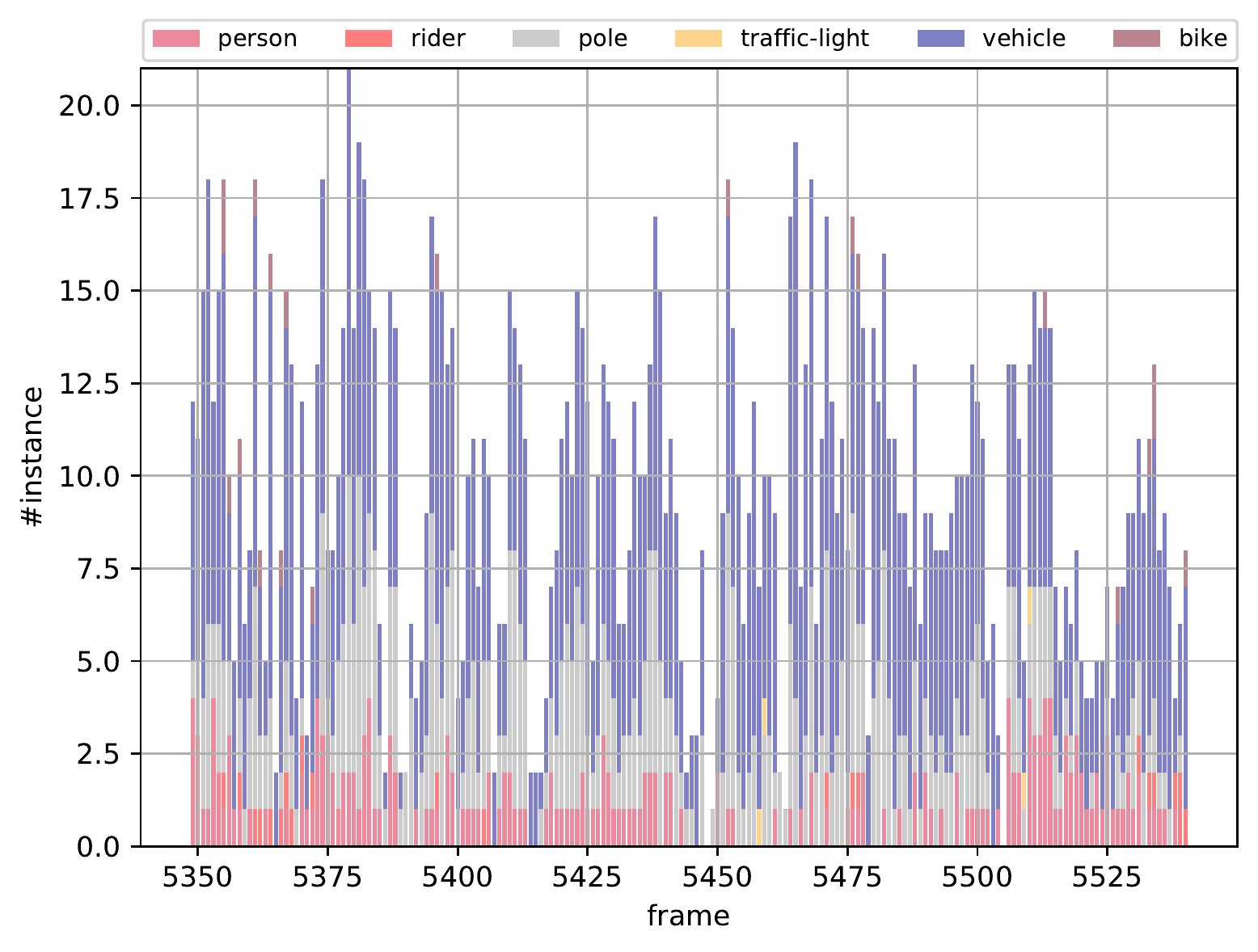}
			\caption[]%
			{{\small Approaching the  Point of Departure}}    
			\label{fig:distr_sub_end}
		\end{subfigure}
		\caption[]
		{\small The spatio-temporal distribution of instances during walking. (a) marks the route along which we acquire the image sequence. (b) shows the distribution of the instances at the initial period of the trip. (c) and (d) serve the same purpose but at different phases of the trip. The rough walking regions corresponding to (b), (c) and (d) are designated in (a) by (b), (c) and (d), respectively. Horizontal axis represents the frame ID whereas the vertical axis denotes the number of occurrences of the instance. } 
		\label{fig:distr_instance}
	\end{figure}

In order to establish that we can fulfill navigational perception needs of visually impaired people by panoptic segmentation in practice, we invite a blindfolded person  wearing the assistive system to navigate around the campus, where the walking route is shown in Figure~\ref{fig:distr_sub_route}. Meanwhile, a sequence of images at the resolution of $480 \times 640$ with $0.25s$ in between each image and the next, which represents the scenes encountered during the navigation of visually impaired pedestrians, is collected by the wearable camera. Figure~\ref{fig:distr_sub_begin} - Figure~\ref{fig:distr_sub_end} present an overview of the spatio-temporal distribution of the instances occurring in front of the blindfolded during the trip. At the beginning of the trip as shown in Figure~\ref{fig:distr_sub_begin}, both of static obstacles like poles and dynamic traffic participants such as vehicles and persons come nearly equally frequent into the observation of the glasses worn by the user. Halfway through the trip as illustrated in Figure~\ref{fig:distr_sub_halfway}, the static obstacle poles dominate the field of view of the wearable camera. Approaching the point of departure as depicted in Figure~\ref{fig:distr_sub_end}, the visually impaired have to be faced with more vehicles. Overall, from the distribution, we can clearly see that the categories of objects met by the blindfolded vary not only from time to time but also from place to place, which means we should consider various objects together in a single framework rather than devise specific algorithms to diverse instance categories. Given that some stuff classes including sidewalk and sky accompany the blindfolded all the way through the trip and their inherent property of being uncountable, we haven't added any stuff information in Figure~\ref{fig:distr_instance}. However, the recognition of the stuff classes like sidewalk is crucial for the safety of visually impaired pedestrians. Both of the above two insights exhibit the necessity of unifying instance and stuff classes segmentation. Our proposed framework acts as the first attempt to cover navigational perception needs for visually impaired people by using panoptic segmentation.
\begin{figure}[t]
    \centering
    \includegraphics[width=\columnwidth]{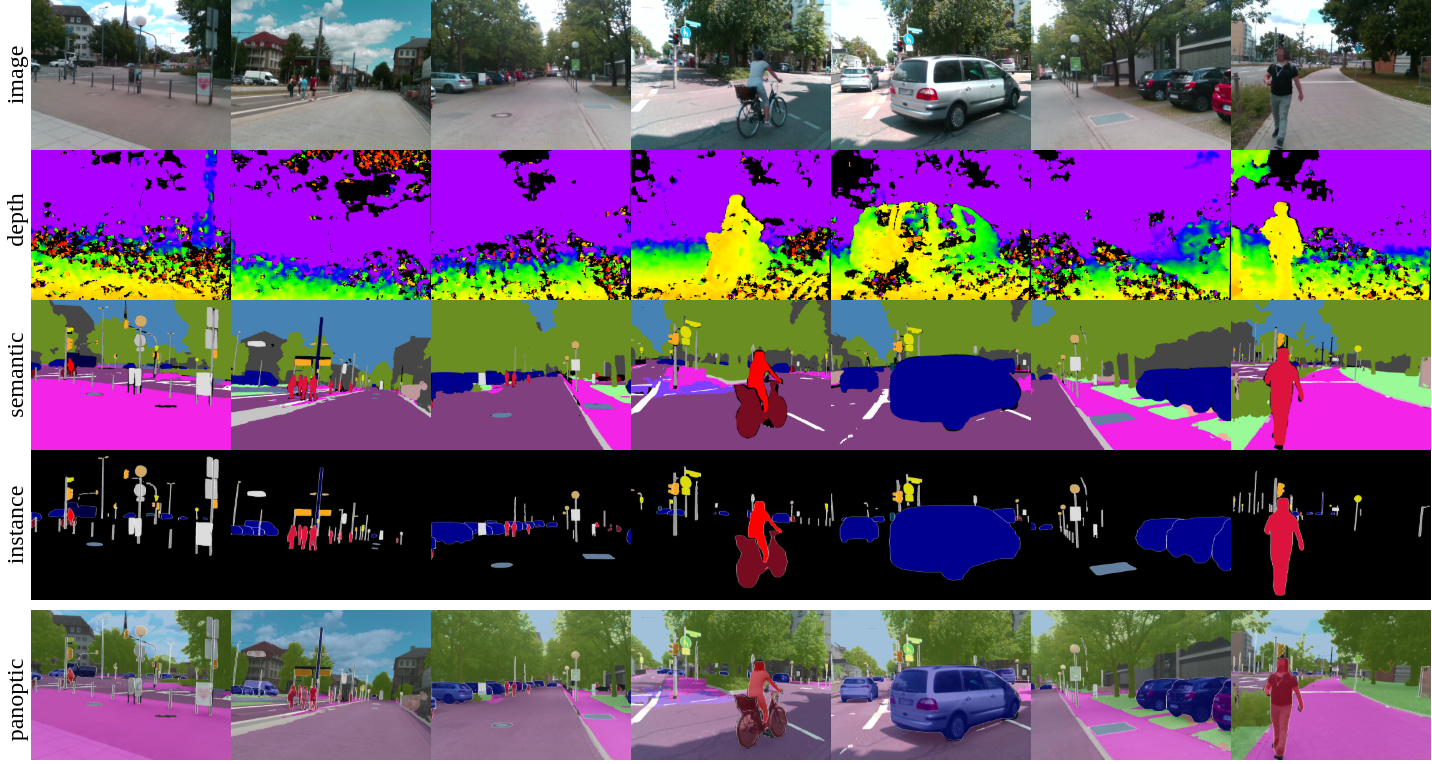}
    \caption{Visualized comparisons between various recognition outputs. Raw RGB image, depth image, semantic segmentation, instance segmentation, panoptic segmentation are organized from top to bottom.}
    \label{fig:qualitative_analysis}
\end{figure}

Various segmentation results are visualized in Figure~\ref{fig:qualitative_analysis}, which evidently showcases the absence of the stuff interpretation in instance segmentation and ignorance of distinction between objects within the same class in semantic segmentation. In contrast, the panoptic segmentation overcomes the fundamental flaws of both instance and semantic segmentation by means of outputting a comprehensive pixel-level set of interpretations, namely per-pixel class label and instance ID. Therefore, the panoptic segmentation serves as an ideal tool for helping the visually impaired to push their environmental perception to a higher level that is both unified and comprehensive. It should be noted from the visualizations that the input at the resolution $480 \times 640$, while not ideal in terms of accuracy, does not hurt the intent of providing a holistic scene understanding too much, because key parts of both stuff and thing classes are precisely and correctly segmented. Additionally, the RGB-D camera presents its versatility by providing depth images, which complement panoptic segmentation and upgrade scene understanding from 2D to 3D. Concretely, our wearable assistive system can tell the visually impaired how far away the objects are in their field of view. 
 \begin{figure}[t]
    	\centering
    \begin{tikzpicture}
    \matrix[matrix of nodes, inner sep=0]{
    \includegraphics[width=0.16\textwidth,keepaspectratio]{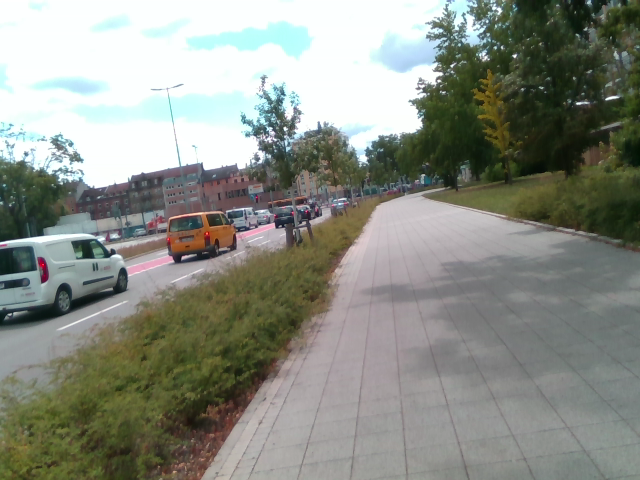} & \includegraphics[width=0.16\textwidth,keepaspectratio]{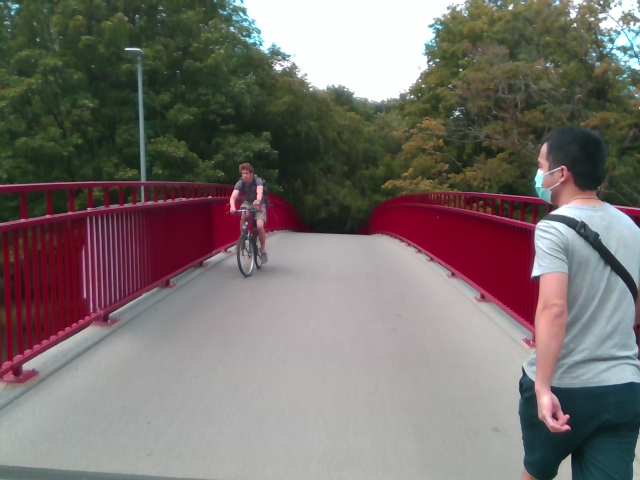} &
    \includegraphics[width=0.16\textwidth,keepaspectratio]{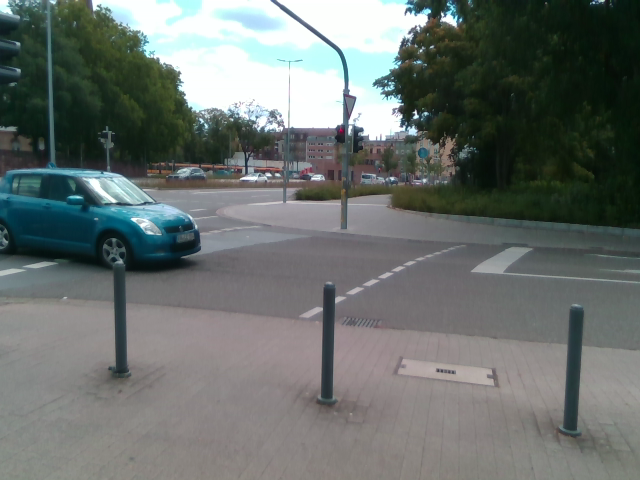} &
    \includegraphics[width=0.16\textwidth,keepaspectratio]{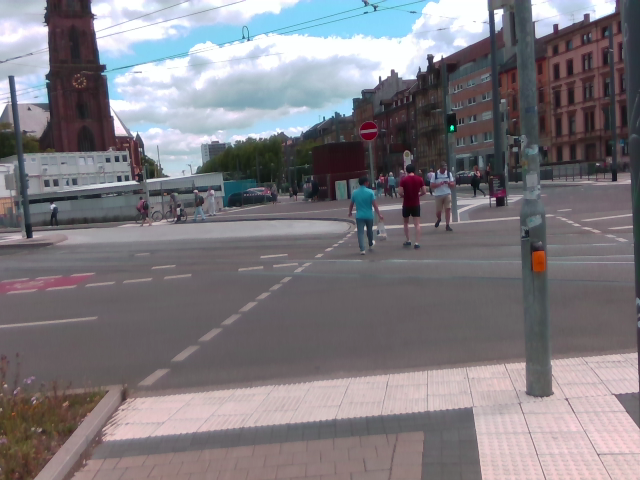} &
    \includegraphics[width=0.16\textwidth,keepaspectratio]{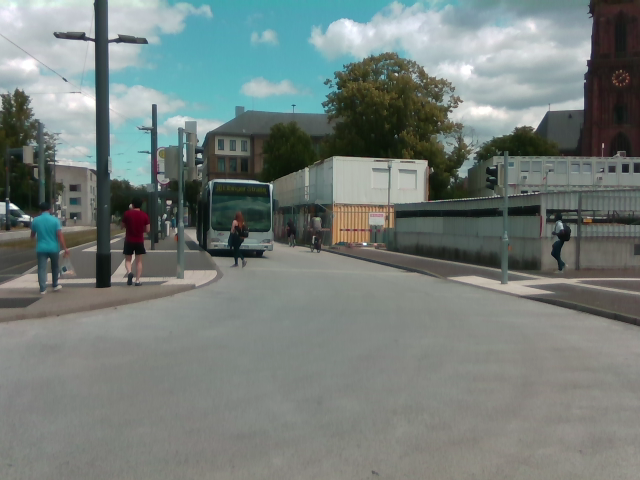} &
    \includegraphics[width=0.16\textwidth,keepaspectratio]{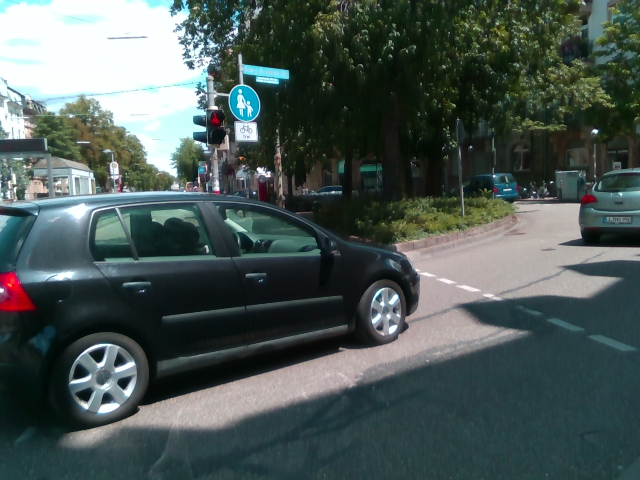} \\
    
    \includegraphics[width=0.16\textwidth,keepaspectratio]{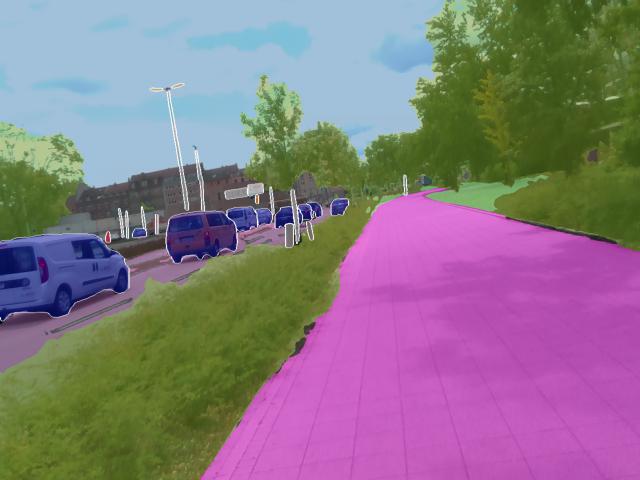} & 
    \includegraphics[width=0.16\textwidth,keepaspectratio]{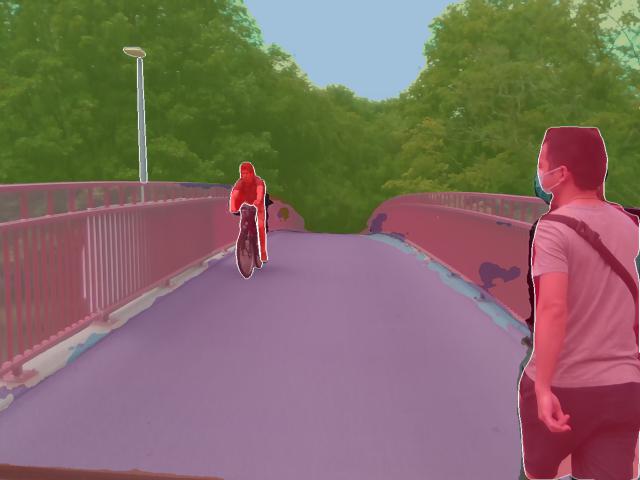} &
    \includegraphics[width=0.16\textwidth,keepaspectratio]{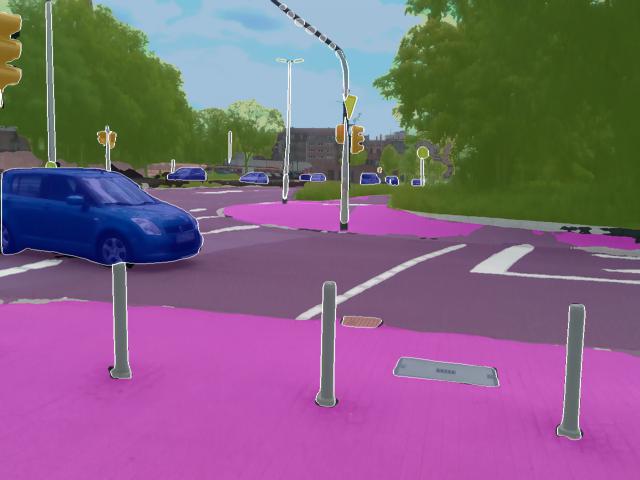} &
    \includegraphics[width=0.16\textwidth,keepaspectratio]{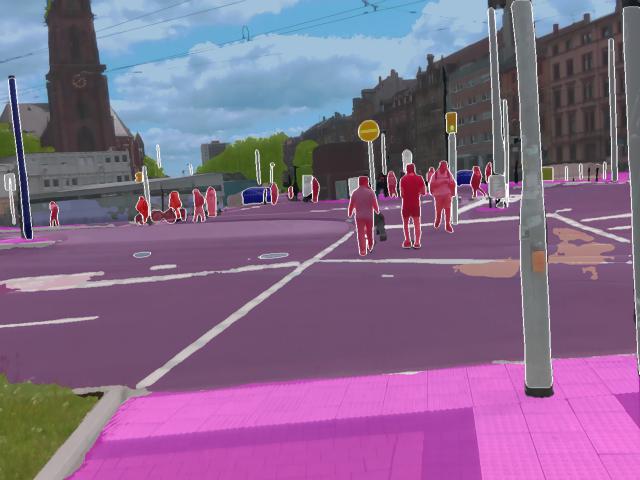} &
    \includegraphics[width=0.16\textwidth,keepaspectratio]{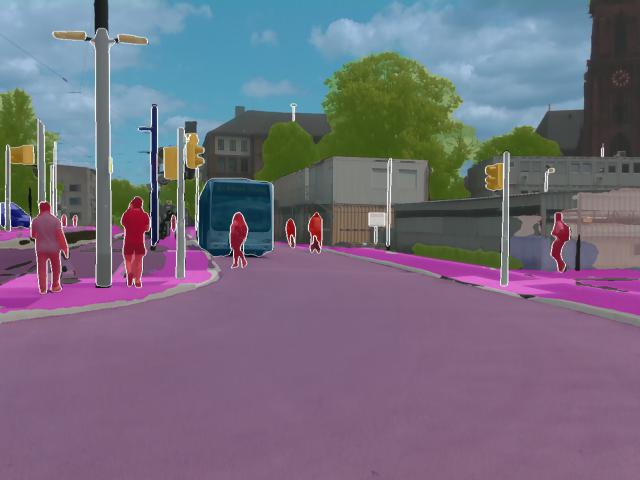} &
    \includegraphics[width=0.16\textwidth,keepaspectratio]{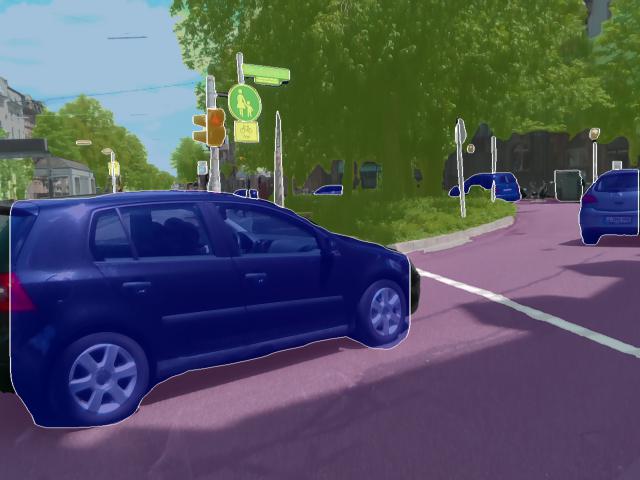} \\

    \phantom{abc} &&&&&\\
    {\scriptsize Sidewalk} & {\scriptsize Bridge} & {\scriptsize  Waiting} & {\scriptsize Crossing Road} & {\scriptsize Bus Station} & {\scriptsize Roundabout} \\
};		  
    \end{tikzpicture}
    \caption{Typical traffic scenes for visually impaired people. Top row: RGB images for the following scenarios: walking on the sidewalk, walking on the bridge, waiting for the traffic light to turn green, entering the bus station and approaching the roundabout. Bottom row: the corresponding panoptic predictions.}
    \label{fig:typical_scenes}
\end{figure}

Furthermore, we extract some typical scenes the visually impaired people usually encounter as illustrated in the Figure \ref{fig:typical_scenes}. As the traffic scenarios range from walking on the sidewalk, crossing the road to approaching the roundabout, the quality of panoptic outputs is  still at a relatively stable and exceptionally remarkable level, which indicates that our system is robust to variations of the environments where the visually impaired are navigating. 
\section{Conclusions}
In this paper, we first present a novel system to satisfy the perception needs by panotpic segmentation that simultaneously addresses the foreground object segmentation and background stuff interpretation on a wearable assistive system. We derive achievability results for various navigational perception tasks by utilizing efficient seamless panoptic segmentation. We gather a sequence of images from the viewpoint of the visually impaired. Based on that, an extensive set of experiments has been conducted, which demonstrates that panoptic segmentation can aid the visually impaired to perceive the surroundings, with semantic and instance-specific interpretation, in a unified and comprehensive way. Our future work will concentrate on 1) developing a more efficient panoptic segmentation architecture, 2) building up a dataset with annotated images taken from the perspective of visually impaired people, and 3) evaluating the novel model on the new dataset.

\section*{Acknowledgement}
This work was supported in part by Federal Ministry of Labor and Social Affairs (BMAS) through the AccessibleMaps project under the grant number 01KM151112, in part by Hangzhou SurImage Technology Company Ltd. and in part by Hangzhou KrVision Technology Company Ltd. (krvision.cn).



%
%
\bibliographystyle{splncs04}
\bibliography{egbib}
\end{document}